\documentclass[letterpaper, 10 pt, journal]{ieeeconf}

\IEEEoverridecommandlockouts

\usepackage{etextools}
\usepackage{amssymb}
\usepackage{float}
\usepackage{makeidx}
\usepackage{amsmath}
\usepackage{bbm}
\usepackage{epsfig}
\usepackage{epsf}
\usepackage{psfrag}
\usepackage{verbatim}
\usepackage{color}
\usepackage{multirow}
\usepackage{tabularx}
\usepackage{booktabs}
\usepackage[tight,footnotesize]{subfigure}
\usepackage{array}
\usepackage{soul}
\usepackage{footnote}
\usepackage{cite}
\usepackage{dblfloatfix}
\usepackage{tcolorbox}
\usepackage{caption}

\usepackage{color, colortbl}
\usepackage[colorlinks,bookmarksnumbered,citecolor=orange,urlcolor=orange]{hyperref}
\usepackage{graphicx}
\graphicspath{{Figures}}
\DeclareGraphicsExtensions{.pdf,.png}
\usepackage{bigstrut}
\usepackage[english]{babel}

\usepackage{csquotes}

\usepackage[T1]{fontenc}
\usepackage{algorithm, setspace}
\usepackage{algpseudocode}
\usepackage{url}
\usepackage{multirow}
\usepackage{float}
\usepackage{xcolor}
\usepackage{hyperref}
 \hypersetup{
     colorlinks=true,
     linkcolor=orange,
     filecolor=orange,
     citecolor=orange,      
     urlcolor=orange,
     }

\usepackage{balance}
\usepackage{amssymb}
\let\labelindent\relax
\usepackage{enumitem}
\usepackage{cuted}

\begin{document}

\title{\LARGE
Safety-Aware Imitation Learning via MPC-Guided Disturbance Injection
}

\author{Le Qiu$^1$, Yusuf Umut Ciftci$^2$, Somil Bansal$^3$ %
\thanks{This work is supported in part by the NSF CAREER Program under award 2240163, the DARPA ANSR program, and the NVIDIA academic grant program.}
\thanks{$^{1}$Le Qiu is with the Department of Electrical Engineering, Tsinghua University, China. {\tt qiule1026@gmail.com}}%
\thanks{$^{2}$Yusuf Umut Ciftci is with the Department of Electrical and Computer Engineering, University of Southern California, USA. {\tt yciftci@usc.edu}}
\thanks{$^{3}$Somil Bansal is with the Department of Aeronautics and Astronautics, Stanford University, USA. {\tt somil@stanford.edu}}%
}

\maketitle

\begin{strip}
\begin{minipage}{\textwidth}\centering
\vspace{-9.2em}
\includegraphics[width=\textwidth]{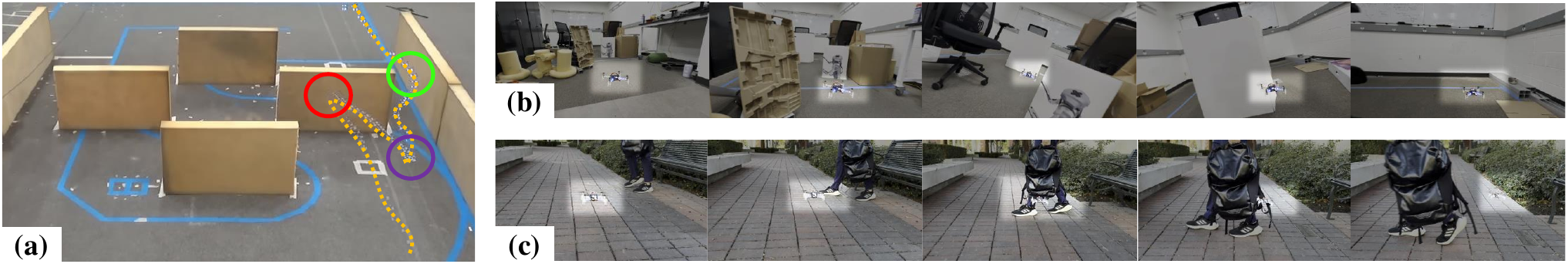}
\captionof{figure}{Our proposed MPC-SafeGIL framework enables safe navigation of a Crazyflie quadrotor across diverse real-world scenarios: (a) in a structured obstacle setting, the quadrotor performs a robust recovery maneuver (\textcolor{orange}{orange})---retreating upon obstacle detection (\textcolor{red}{red}), reorienting to adjust heading (\textcolor{violet}{purple}), and ultimately exiting through an available opening (\textcolor{green}{green}); (b) in densely cluttered environments; and (c) in dynamic scenarios with moving obstacles.}
\label{fig:front figure}
\vspace{-1.2em}
\end{minipage}
\end{strip}

\begin{abstract}
Imitation Learning has provided a promising approach to learning complex robot behaviors from expert demonstrations. 
However, learned policies can make errors that lead to safety violations, which limits their deployment in safety-critical applications. 
We propose MPC-SafeGIL, a design-time approach that enhances the safety of imitation learning by injecting adversarial disturbances during expert demonstrations. 
This exposes the expert to a broader range of safety-critical scenarios and allows the imitation policy to learn robust recovery behaviors.
Our method uses sampling-based Model Predictive Control (MPC) to approximate worst-case disturbances, making it scalable to high-dimensional and black-box dynamical systems.
In contrast to prior work that relies on analytical models or interactive experts, MPC-SafeGIL integrates safety considerations directly into data collection. 
We validate our approach through extensive simulations including quadruped locomotion and visuomotor navigation and real-world experiments on a quadrotor, demonstrating improvements in both safety and task performance. See our website here: \url{https://leqiu2003.github.io/MPCSafeGIL/}

\end{abstract}

\section{Introduction}

Imitation learning has demonstrated great potential to enable robots to learn complex robotic behaviors from expert demonstrations. Behavior Cloning (BC), which maps states or observations directly to actions by emulating expert behavior, is a fundamental technique due to its simplicity and wide applicability. It has been applied across a variety of robotic domains, including manipulation tasks \cite{finn2017one, zhao2023learning}, legged locomotion \cite{bin2020learning}, and autonomous navigation \cite{condinional_il, agile_auto_driving}. 
However, discrepancies between the expert and the learned imitation policy can lead to safety violations, particularly in safety-critical systems, due to potential learning errors, external disturbances, or covariate shift \cite{compounding_error_covariate_shift, dagger}. 

Methods that address the safety challenges in imitation learning can be broadly categorized into deployment-time (or \textit{test-time}) and design-time (or \textit{training-time}) strategies. 
Among deployment-time solutions, safety filter has become a popular technique to verify the safety of learned policies and intervene when safety constraints are at risk of being violated \cite{choi2021robust, wabersich2023data, hsu2023safety, yang2024enhancing}. Various methods have been developed to construct safety filters, including Control Barrier functions (CBF) \cite{ames2019control, prajna2004safety}, HJ reachability \cite{bajcsy2019efficient}, predictive rollout-based simulations \cite{nguyen2024gameplay}, and learning from demonstrations\cite{yang2024enhancing}. Despite their strong theoretical guarantees, deploying safety filters in the context of imitation learning presents several limitations. First, constructing and updating online filters remains difficult in autonomous systems, particularly when full state information is not directly available (e.g., robots operating on vision-based systems) and in high-dimensional dynamics where real-time computation is nontrivial. Second, the overall task performance of test-time safety filters is inherently limited by the quality and safety-awareness of the underlying learned policy.

These limitations highlight the need to \textit{design} imitation policy that proactively embeds safety considerations into the learned policy. 
Design-time methods primarily aim to reduce learning error and mitigate covariate shift, which in turn can lower the likelihood of unsafe behaviors (assuming the expert demonstrations themselves are safe). Off-policy approaches achieve this by generating additional corrective action labels to augment expert demonstrations \cite{ke2023ccil, zhou2023nerf, hoque2024intervengen}.
On-policy approaches iteratively roll out the current policy and query expert feedback to reduce compounding errors \cite{laskey2017dart, dagger}. 
Adversarial imitation learning approaches \cite{gail, airl, sqil} refine policies by learning to distinguish expert from the learned agent through a discriminator. Other methods augment standard BC losses to improve robustness and stability around demonstrations states \cite{tasil, stableBC}.
However, these approaches do not explicitly model safety as a first-class objective during data collection or policy optimization. This lack of intentional safety modeling means that imitation policies may still fail in rare but high-risk scenarios that were not adequately represented in training data.

A recent approach, SafeGIL\cite{ciftci2024safe}, introduces a design-time method to overcome this gap by injecting optimal adversarial disturbances into expert demonstrations. By guiding the expert toward safety-critical states, this strategy allows the learned policy to observe and imitate recovery behaviors in risky states. However, SafeGIL relies on Hamilton-Jacobi (HJ) reachability analysis \cite{bansal2017hamilton, mitchell2005time, mitchell2004toolbox}, which suffers from the curse of dimensionality and does not scale beyond 5-6 state dimensions \cite{bansal2017hamilton}. Additionally, it requires an analytical expression of robotic dynamics, limiting its applicability to black-box or other general robotic systems.

To address the aforementioned challenges, we propose MPC-SafeGIL, a scalable design-time framework for safety-guided imitation learning in high-dimensional and black-box dynamical systems. Our \textit{key insight} is to leverage sampling-based MPC, specifically the Model Predictive Path Integral (MPPI) algorithm \cite{aggressivemppi, Borquez2025DualGuard, multirotor}, to approximately synthesize optimal adversarial disturbance that simulates worst-case safety-critical deviations at each state. These disturbances emulate potential errors the policy may encounter at test time. By injecting such adversarial disturbances into expert demonstrations, we expose the system to a wider distribution of meaningful critical states, thereby enabling the imitation policy to learn robust recovery strategies. Our method is designed to be highly parallelizable on GPUs, allowing for efficient trajectory sampling. We validate MPC-SafeGIL through extensive experiments in both simulation and real-world robotic platform. In summary, our contributions are: 
\begin{itemize}
    \item \textbf{Safety-Aware Imitation Learning:} We propose MPC-SafeGIL, a framework that proactively steers expert demonstrations toward safety-critical regions using sampling-based MPC, resulting in safer learned policies.

    \item \textbf{Scalability to High-Dimensional and Black-Box Dynamical Systems:} By leveraging sampling-based optimization, MPC-SafeGIL scales to high-dimensional systems without requiring analytical dynamic models. We validate the performance in both simulation and hardware experiments.

    \item \textbf{Complementary to Online Safety Mechanisms:} We demonstrate that MPC-SafeGIL can be seamlessly integrated with deployment-time safety mechanisms, such as safety filter, to further improve safety and task performance during deployment.
\end{itemize}

\section{Problem Formulation} \label{sec:problem}
We consider a robotic system with discrete-time dynamics $x_{t+1} = f(x_t, u_t)$, where $x \in \mathcal{X} \subseteq \mathbb{R}^{n_x}$ is the state and $u \in \mathcal{U} \subseteq \mathbb{R}^{n_u}$ is the control. We assume that the dynamics function $f$ can be accessed as a black box (e.g., through a simulator).
Safety is encoded through a failure set $\mathcal{L}$, defined as the sub-zero level set of a function $l: \mathbb{R}^{n_x} \to \mathbb{R}$, i.e., $\mathcal{L} = \{x: l(x) \leq 0\}$. For example, in navigation, $\mathcal{L}$ may represent obstacle regions, and $l(x)$ could be the signed distance from the robot to those obstacles.

Our objective is to learn a safe imitation policy $\pi_\theta(\cdot)$, parameterized by $\theta \in \Theta$, that imitates a given expert policy $\pi^*(\cdot)$ while minimizing safety violations. Specifically, the learned policy should keep the system outside the failure set $\mathcal{L}$ during test time, without sacrificing task performance.

\section{MPC-Based Safety Guidance} \label{sec:method}

\subsection{Injection of Adversarial Disturbance}
Our approach (summarized in Algorithm~\ref{our_algorithm}) aims to learn a safe imitation policy by intentionally guiding expert demonstrations toward safety-critical states. To achieve this, we abstract potential test-time errors as adversarial disturbances $d(x)$ that push the system toward unsafe regions. The objective is to compute the optimal adversarial disturbance $d^*(x)$ and inject it into the expert policy $\pi^*(\cdot)$ during data collection, forming a guided expert policy $\pi^G(\cdot)$:
\begin{equation}
    \pi^G(x) = \pi^*(x) + d^*(x).
\end{equation}
Ideally, $d^*(x)$ simulates the most safety-critical deviations that may occur during deployment. By perturbing the expert in this way, the system encounters safety-critical scenarios more frequently. The demonstrations collected under $\pi^G(\cdot)$ are then used to train an imitation policy $\pi_\theta(\cdot)$. The corrective actions of the expert from safety critical states in the training dataset enable the imitation policy to learn effective recovery behaviors and ensure reliable performance during test time.

\begin{figure}[t]
\begin{algorithm}[H]
    \caption{Safety Guided Data Collection}
    \label{our_algorithm}
    \begin{algorithmic}
        \State \textbf{Input:} $\pi^*(\cdot)$, $\bar{d}_{max}$, Number of demonstrations $K$, Trajectory length $T$
        \State \textbf{Initialize:} $\mathcal{D} \gets \emptyset$
        \For{$k=1:K$}
            \State Sample initial state $x_1$
            \For{$t=1:T$}
                \State $\bar{d} \sim U(0,\bar{d}_{max})$
                \State Get optimal disturbance $d^*(x_t; \bar{d})$ using \eqref{eq:opt_dist}
                \State $\pi^G(x_t) \gets \pi^*(x_t) + d^*(x_t; \bar{d})$
                \State $\mathcal{D} \gets \mathcal{D} \cup (x_t, \pi^*(x_t))$
                \State $x_{t+1}=f(x_t,\pi^G(x_t))$
            \EndFor
        \EndFor
    \end{algorithmic}
\end{algorithm}
\vspace{-1.5em}
\end{figure}

\subsection{Computation of Adversarial Disturbance}
We draw inspiration from HJ reachability analysis to compute the adversarial disturbance. HJ reachability typically determines the backward reachable tube (BRT) of a system, the set of all states from which no matter what control policy $u(\cdot)$ does, there exists a disturbance strategy $d(\cdot)$ that will drive the system to the failure set $\mathcal{L}$. To formalize this, we define a cost function as the minimal ``distance'' to the failure set along a robot trajectory starting at $x$ and governed by $u(\cdot)$ over the time horizon $t,t+1,\dots,T$:
\begin{equation}
    J(x_t, u(\cdot), d(\cdot)) = \min_{k \in \{t,\dots, T\}} l(x_k).
\end{equation}

\noindent \textbf{Disturbance Computation via Robust Optimal Control.}
We formulate the problem as a two-player zero-sum dynamic game. The goal is to find a control sequence $u(x)$ that prevents the system from entering the failure set $\mathcal{L}$, despite the worst-case disturbance. Thus, we want to find the control strategy $u(\cdot)$ that maximizes the worst-case cost function, when the disturbance $d(\cdot)$ tries to minimize it. The disturbance perturbs the control input directly, leading to the dynamics of the form $x_{t+1}=f(x_t, u_t+d_t)$.
To simplify the formulation, We assume symmetric input bounds: $u \in [-\bar{u}, \bar{u}]$ and $d \in [-\bar{d}, \bar{d}]$. While this assumption facilitates mathematical tractability, it is not strictly necessary and can be relaxed in practice. The interaction between control and disturbance establishes a max-min problem:
\begin{equation}
    \begin{gathered}
        \max_{u(\cdot)} \min_{d(\cdot)} J(x_t, u(\cdot), d(\cdot)), \\
        x_{k+1}=f(x_k, u_k+d_k), \quad \forall k \in \{t, t+1, \dots, T\},\\
        |u_k| \leq \bar{u}, \quad
        |d_k| \leq \bar{d}.
    \end{gathered}
\end{equation}

\noindent \textbf{Single-Player Optimal Control.}
Solving the two-player robust optimal control problem is computationally challenging, especially in high-dimensional systems. To address this, we reformulate the problem as a single-player optimal control problem under the assumption of control-affine dynamics. Specifically, we combine control and disturbance into a single input, $w=u+d$, with an adjusted bound $|w| \leq \bar{u} - \bar{d}$. This converts the original max-min formulation into a single max problem:
\vspace{-0.5em}
\begin{equation}
    \begin{gathered}
        \max_{w(\cdot)} J(x_t, w(\cdot)), \\
        x_{k+1}=f(x_k,w_k), \quad \forall k \in \{t, t+1, \dots, T\},\\
        |w_k| \leq \bar{u} - \bar{d}.
    \end{gathered}
    \label{eq:single_player}
\end{equation}

A formal proof for this reduction with control-affine dynamics is provided in Appendix. A wide range of robotic systems can be modeled as control-affine \cite{lynch2017modern}. Even for non-affine dynamics, our method provides a useful safety-guidance approximation, as demonstrated in our case studies. Since non-affine systems can often be linearized via first-order Taylor expansion, the control-affine assumption is rarely restrictive in practice. This reduction significantly lowers the computational burden while retaining the ability to capture adversarial effects, enabling practical application in high-dimensional safety-critical scenarios.

\noindent \textbf{MPPI Solution for Disturbance.} After reformulating the problem as a single-player optimal control problem, it can be solved effectively using sampling-based MPC algorithms. Specifically, we leverage MPPI to find the optimal adversarial disturbance. First, we generate $N$ control sequences, denoted by $\mathcal{W} = \{\mathbf{w_0}, \mathbf{w_1},...,\mathbf{w_N}\}$, each of length $H$. We simulate the system dynamics in parallel and evaluate the cost function $J(x,\mathbf{w_0}),...,J(x,\mathbf{w_N})$. This sampling-based approach is computationally efficient and well-suited for high-dimensional and black-box dynamical systems. Next, the top $K$ control sequences with highest costs are selected:
\[
\{\mathbf{w_1}, \mathbf{w_2},...,\mathbf{w_K}\} = \operatorname{arg\,top}_K \{J(x,\mathbf{w_{i}}) \mid \mathbf{w_{i}} \in \mathcal{W}\},
\]
Finally, the control sequence is updated via a weighted sum of the selected trajectories using a temperature parameter $\lambda$:
\begin{equation}
    \label{eq:mppi_optimal_control}
    \mathbf{w^*}(x) = 
    \sum_{k=1}^{K}
    \left(
    \frac{\exp\left(-\frac{1}{\lambda} J(x,\mathbf{w_{k}})\right) \mathbf{w_{k}}}
    {\sum_{k=1}^{K} \exp\left(-\frac{1}{\lambda} J(x,\mathbf{w_{k}})\right)}
    \right).
\end{equation}

After obtaining the optimal control sequence $\mathbf{w^*}(x)$, we extract its first control input $w^*(x)$ and recover the corresponding disturbance under the control-affine assumption. This yields the solution to the original two-player robust control problem:
\vspace{-0.5em}
\begin{equation}
    d^*(x) = 
    \begin{cases}
        \bar{d} & \text{if } w^*(x) < 0, \\
        -\bar{d} & \text{if } w^*(x) > 0.
    \end{cases}
    \label{eq:opt_dist}
\end{equation}
which is then injected into expert policy $\pi^*(x)$ to deliberately guide the system toward safety-critical states. This guided dataset is used to train the imitation policy $\pi_{\theta}(x)$.

\section{Experiments} \label{sec:experiments}
In this section, we demonstrate the safety and performance enhancement achieved by MPC-SafeGIL in high-dimensional and black-box dynamical systems. We consider three case studies: quadruped navigation with LiDAR-based sensing, autonomous racing car with LiDAR-based perception, and quadrotor navigation in real-world environments.

\subsection{Verification of Adversarial Disturbance} 
\begin{figure}[h]
    \centering
\includegraphics[width=\linewidth]{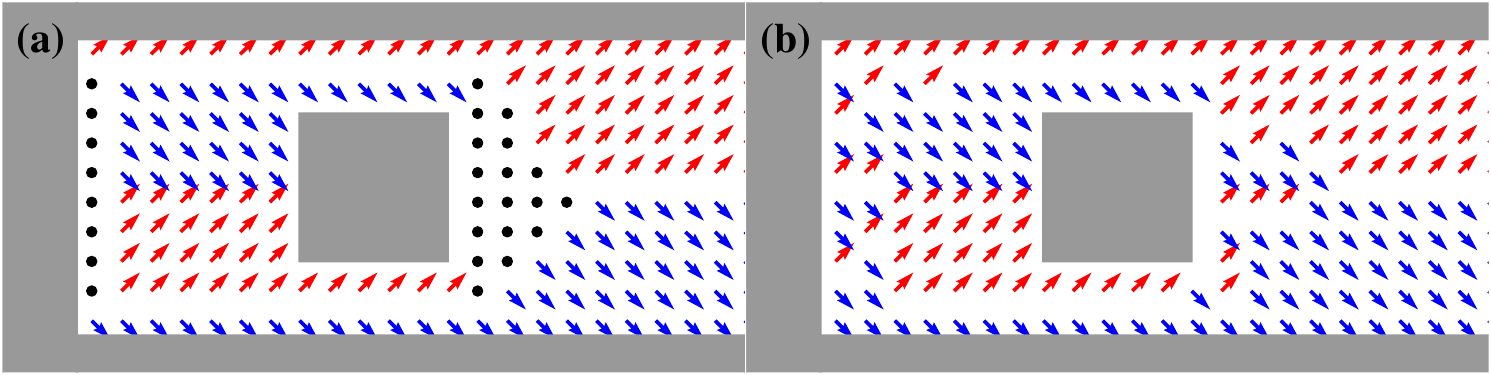}
    \caption{The optimal adversarial disturbance on the state slice $(p_x, p_y, 0)$. The gray areas are obstacles. Red and blue arrows indicate different disturbance directions. (a) Disturbance computed via HJ reachability. Black dots mark states where the spatial gradient of value function is zero. (b) Disturbance computed via MPPI. Note the similarity of the disturbance fields.}
    \label{fig:Dubins_verification}
    \vspace{-1.5em}
\end{figure}

We first verify that the optimal adversarial disturbance computed by our MPPI-based method is consistent with the one derived from HJ reachability \cite{ciftci2024safe}. We consider a 3D Dubins car system tasked with reaching a goal without colliding with obstacles. The state is given by $x=(p_x, p_y, \theta)$, where $(p_x,p_y)$ is the robot's 2D position and $\theta$ is the robot's heading. The control input is yaw velocity $\omega$, bounded by $|\omega| \leq \bar{\omega}$. The system dynamics are: 
\begin{equation}
    \label{eq:reduced-order}
    \begin{aligned}
        p_{x,t+1} &= p_{x,t} + \delta \cdot \cos \theta, \\
        p_{y,t+1} &= p_{y,t} + \delta \cdot \sin \theta, \\
        \theta_{t+1} &= \theta_{t} + \delta \cdot \omega.
    \end{aligned}
\end{equation}
We set the discretization step $\delta = 0.1 \mathrm{s}$, $ \bar{\omega} = 1 \mathrm{rad/s}$, $v = 1 \mathrm{m/s}$, and $\bar{d}=0.5\bar{\omega}$.
The HJ reachability value function is computed on a $101 \times 101 \times 101$ grid using the Level Set Toolbox \cite{mitchell2004toolbox}. The converged value function is then used to derived the optimal disturbance $d^*$. The disturbance field on the state slice $(p_x, p_y, 0)$ is shown in Fig.~\ref{fig:Dubins_verification}(a). Different colors indicate different directions. Black dots represent states where the spatial derivative of the reachability value function is zero, indicating that the disturbance direction has no effect on the safety in these regions. Fig.~\ref{fig:Dubins_verification}(b) shows the $d^*$ computed by MPPI, where we use $N=1000$ samples, an optimization horizon of $H=20 (2s)$, and a smooth parameter $K=10$. Since MPPI does not rely on explicit gradients of value function, zero-gradient regions are not identified. The two disturbance fields are largely consistent. Quantitatively, we sample 100{,}000 random states in the blank space where the spatial gradient is non zero. We observe a low mean squared error (MSE) of 0.19 for the optimal disturbance between the two methods, demonstrating that the MPPI-based method successfully captures the same adversarial behavior as HJ reachability,validating the proposed approach to compute the adversarial disturbance.

\subsection{Quadruped Navigation through Cluttered Environment}
To demonstrate the robustness of MPC-SafeGIL in high-dimensional systems with black-box dynamics, we consider a quadruped navigation task. The objective is to reach a goal position while avoiding obstacles, using onboard LiDAR and proprioception. The policy is trained in randomly generated environments with varying obstacle configurations and tested in an unseen setting with randomized starting positions. We use Isaac Gym \cite{makoviychuk2021isaac} to enable massively parallel simulations on GPUs with black-box dynamics. The quadruped robot is equipped with a LiDAR sensor that emits $54$ evenly spaced rays over a $270^\circ$ field of view. The control is hierarchical: a high-level policy (to be learned via imitation) outputs linear and yaw velocity commands $[v, w] \in [-0.5, 1.5] \mathrm{m/s} \times [-1.0, 1.0] \mathrm{rad/s}$, and a low-level RL-based policy \cite{margolis2022walktheseways} tracks these commands through joint-level actuators. 

Training demonstrations are collected in randomly generated maps. Each environment contains four cylinder obstacles, with centers sampled uniformly from $[x,y] \in [2.5, 7.5]\mathrm{m} \times [-2, 2]\mathrm{m}$ and the radius sampled uniformly from $[0.1,1.0] \mathrm{m}$. During demonstration, the quadruped navigates from $(0, 0)$ to the goal $(10, 0)$. The failure set $\mathcal{L}$ is defined as the union of obstacle regions, and $l(x)$ is the signed distance function to the nearest obstacle boundary. The expert is a sampling-based high-level MPC planner that minimizes distance-to-goal and penalizes collisions. It operates on a reduced-order state consisting of position and yaw angle $(p_x, p_y, \theta)$ and reduced-order dynamics in \eqref{eq:reduced-order}.
The imitation policy takes as input a 92D observation $x=(o, p_{rel}, v_x, v_y, \dot{\theta}, q, \dot{q}, g_p, c)$, which includes LiDAR readings $o$, relative goal position in the robot's frame $p_{rel}$, COM linear and angular velocities $(v_x, v_y, \dot{\theta})$, joint positions and velocities $q$, projected gravity $g_p$, and foot contacts $c$. It is modeled as an MLP that predicts the high-level commands $(v, w)$, and is trained with 10 different random seeds to capture variance.

\begin{figure}[t]
    \centering
    \includegraphics[width=\linewidth]{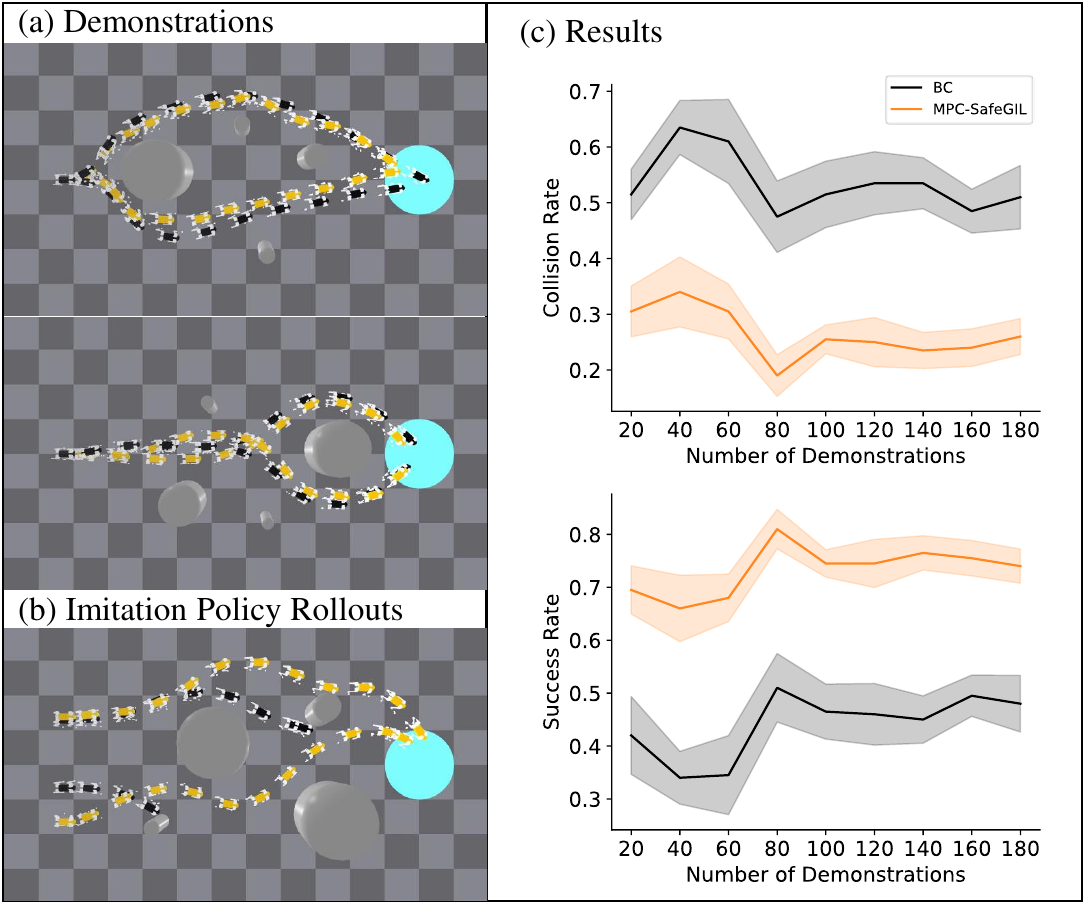}
    \caption{Comparison between BC (\textcolor{black}{black}) and MPC-SafeGIL (\textcolor{orange}{orange}). (a) Demonstrations collected in different environments navigating to the goal (\textcolor{cyan}{cyan}). (b) Rollouts of policies learned from 80 demonstrations. (c) Collision rate and success rate across the number of demonstrations. MPC-SafeGIL achieves a significant safety improvement.}
    \label{fig:Quadruped_Safety}
    \vspace{-1.5em}
\end{figure}

Evaluation is conducted in an unseen environment to assess generalization, where the goal is fixed at $(10,0)$ and the robot starts at $x=0$ with the $y$-coordinate randomly sampled from $[-2,2]$. Each policy is evaluated over 20 rollouts. Safety is measured by collision rates (proportion of trajectories that collide before reaching the goal) and performance is measured by success rates (proportion of trajectories that reach the goal without collisions).

MPC-SafeGIL injects adversarial disturbance with bound $\bar{d}_{max}=0.7\,\bar{u}$, allowing up to 70\% imitation error. To optimize the adversarial disturbance using \eqref{eq:opt_dist}, we use $N=500$ samples, a planning horizon of $H=200 (4s)$, and a smooth parameter $K=10$. Unlike the expert's reduced-order model, MPC-SafeGIL uses full-order dynamics during disturbance optimization in order to produce more realistic and informative adversarial perturbations. Disturbances are applied to the high-level command space $[v, w]$ to generate noise-injected demonstrations. 

\subsubsection*{\textbf{Safety}} We compare MPC-SafeGIL against vanilla Behavior Cloning (BC) across different numbers of demonstrations. The results are shown in Fig.~\ref{fig:Quadruped_Safety}(c). MPC-SafeGIL consistently achieves lower collision rates, especially in low-data regimes where imitation errors are more likely and safety is critical. Notably, with only 80 demonstrations, MPC-SafeGIL reduces the collision rate to approximately 25\%. As shown in Fig.~\ref{fig:Quadruped_Safety}(a), MPC-SafeGIL disturbances consistently guide the expert closer to obstacles in different environments, enabling the policy to observe and learn from safety-critical scenarios. Consequently, the learned policy exhibits stronger recovery behaviors and generalizes better during testing, such as executing tight turns near obstacles (Fig.~\ref{fig:Quadruped_Safety}(b)). In contrast, since the expert planner is inherently safe and conservative, even with more collected trajectories, it may fail to encounter challenging near-collision states, leading to slower safety improvements in standard BC.

It is also worth noting that SafeGIL \cite{ciftci2024safe} is not directly applicable to this case study due to the computational challenges associated with high dimensionality and the black-box nature of the system dynamics. Additionally, the random generation of environments poses a challenge, as it requires solving the expensive reachability problem for each instance.

\begin{figure}[h]
    \centering
\includegraphics[width=\linewidth]{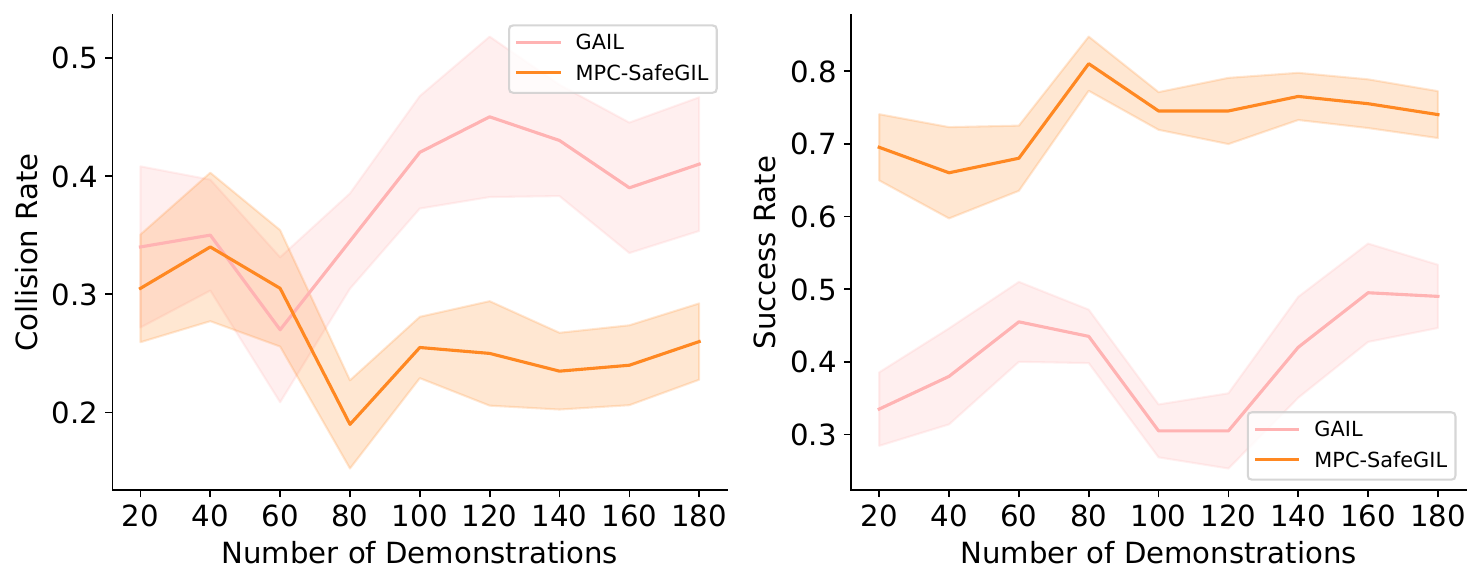}
    \caption{Comparison between MPC-SafeGIL and GAIL. MPC-SafeGIL consistently achieves higher success rates, even when both method exhibit comparable collision rates.}
    \vspace{-1.5em}
    \label{fig:Quadruped_Tradeoff}
\end{figure}

\subsubsection*{\textbf{Performance}} We compare MPC-SafeGIL against \textbf{GAIL}\cite{gail}, a generative adversarial imitation learning method using a learned discriminator as reward. As shown in Fig.~\ref{fig:Quadruped_Tradeoff}, although GAIL achieves a comparable collision rate to MPC-SafeGIL with fewer than 60 demonstrations, its success rate remains substantially lower. This is because, with limited data, GAIL tends to generate overly conservative trajectories that avoid obstacles but often fail to reach the goal. Typical failure modes include circling around obstacles or missing the goal even when nearby. These issues arise from GAIL's reliance on local state-action reward signal, which limits long-horizon goal-reaching behaviors. With more demonstrations, GAIL begins to imitate the expert's direct behaviors through cluttered areas, improving path optimality but increasing collision rates due to the lack of explicit collision penalty. This highlights GAIL's difficulty in balancing safety and performance without structured guidance. In contrast, MPC-SafeGIL explicitly preserves both safety and goal-reaching objectives by injecting structured disturbances during demonstration. This trend is also consistent with the success rate gap observed between BC and MPC-SafeGIL in Fig.~\ref{fig:Quadruped_Safety}(c), where MPC-SafeGIL reduces collision rates while improving overall success rates. These results demonstrate that MPC-SafeGIL effectively resolves the conventional trade-off between safety and task performance.

\begin{figure}[h]
    \centering
    \includegraphics[width=\linewidth]{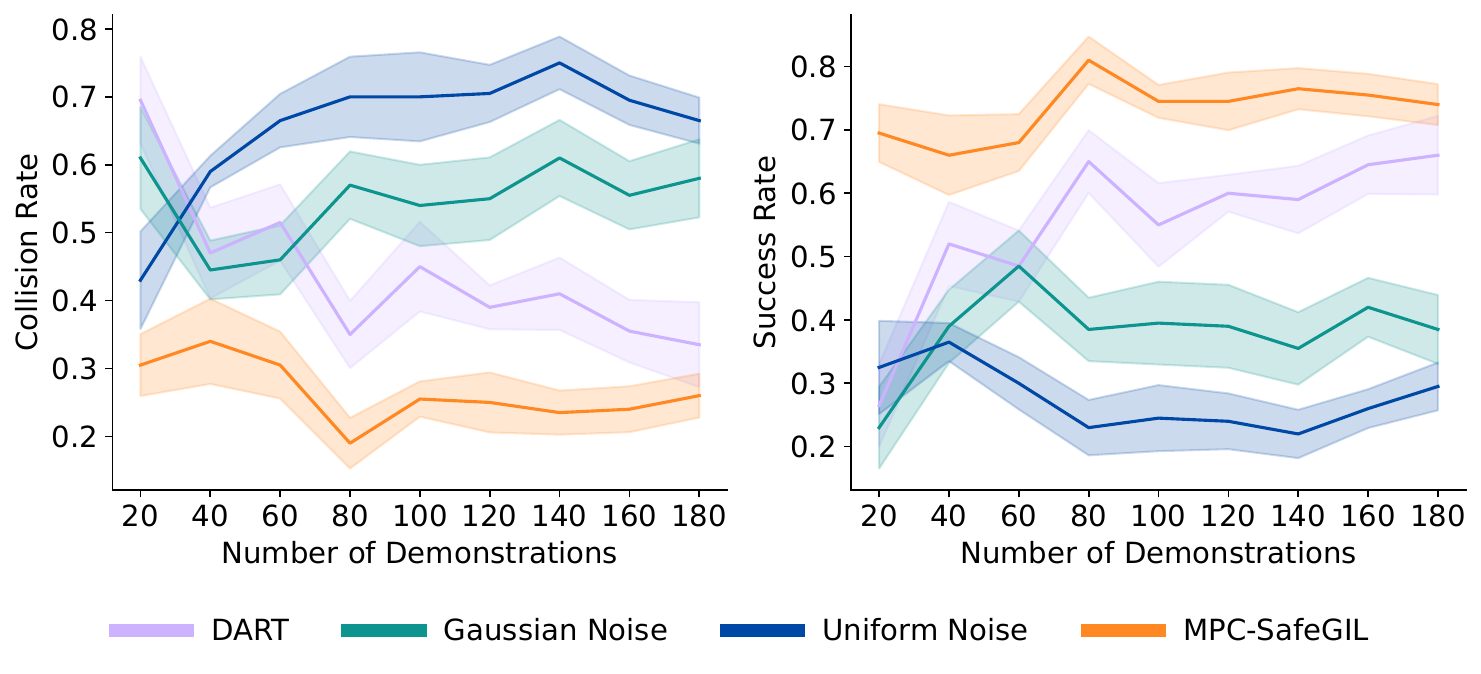}
    \caption{Comparison of different noise-injection schemes. MPC-SafeGIL achieves the lowest collision rate and the highest success rate.}
    \vspace{-1.0em}
    \label{fig:Quadruped_Noise}
\end{figure}

\subsubsection*{\textbf{Adversarial Disturbance}} MPC-SafeGIL can be considered as a data augmentation scheme that inject adversarial noise to expert demonstrations. We compare its effectiveness against three alternative noise-injection baselines: \textbf{Gaussian Noise BC}, \textbf{Uniform Noise BC}, and \textbf{DART}\cite{laskey2017dart}, which uses online noise covariance estimation. Fig.~\ref{fig:Quadruped_Noise} shows that injecting random noise (Gaussian/Uniform) fails to improve either safety or performance. DART yields some improvements by adaptively estimating the noise covariance based on the discrepancy between the learned policy and the expert. However, MPC-SafeGIL achieves the lowest collision rate and the highest success rate. This advantage stems from its adversarial formulation: instead of random perturbations, MPC-SafeGIL purposely targets greater exploration of safety-critical states. While DART focuses on improving imitation fidelity, its safety benefits are a second-order effect. MPC-SafeGIL, on the other hand, explicitly prioritizes safety during data collection, demonstrating the value of adversarial disturbance in improving both safety and task performance.

\begin{table}[h]
    \centering
    \begin{tabular}{cccc}
\toprule
$\bar{d}_{max} / \bar{u}$ &  $0.3$ & $0.5$ & $0.7$ \\
\midrule
Collision rate & $0.37_{(0.05)}$ & $0.34_{(0.06)}$ & $\mathbf{0.19}_{(0.04)}$ \\
\bottomrule
\end{tabular}

    \caption{Mean collision rates of policies trained with MPC-SafeGIL under different disturbance bounds.}
    \vspace{-1.0em}
    \label{tab:disturbance_bounds}
\end{table}

\subsubsection*{\textbf{Effect of Disturbance Bound}} To understand how disturbance bounds affect the expert's exposure to safety-critical states, we conduct experiments with different relative disturbance bounds $\bar{d}_{max} / \bar{u}$ (Table.~\ref{tab:disturbance_bounds}). As the bound increases, stronger adversarial disturbance perturbations drive the expert into a broader range of risky states, yielding demonstrations with rich corrective behaviors that improve policy robustness and reduce collisions. However, when the disturbance bound exceeds a certain point, the robot may be pushed into unrecoverable states, making demonstrations less informative and ultimately harming policy performance.
\vspace{-1.5em}

\subsection{F1Tenth}

\begin{figure}[h]
    \centering
    \includegraphics[width=\linewidth]{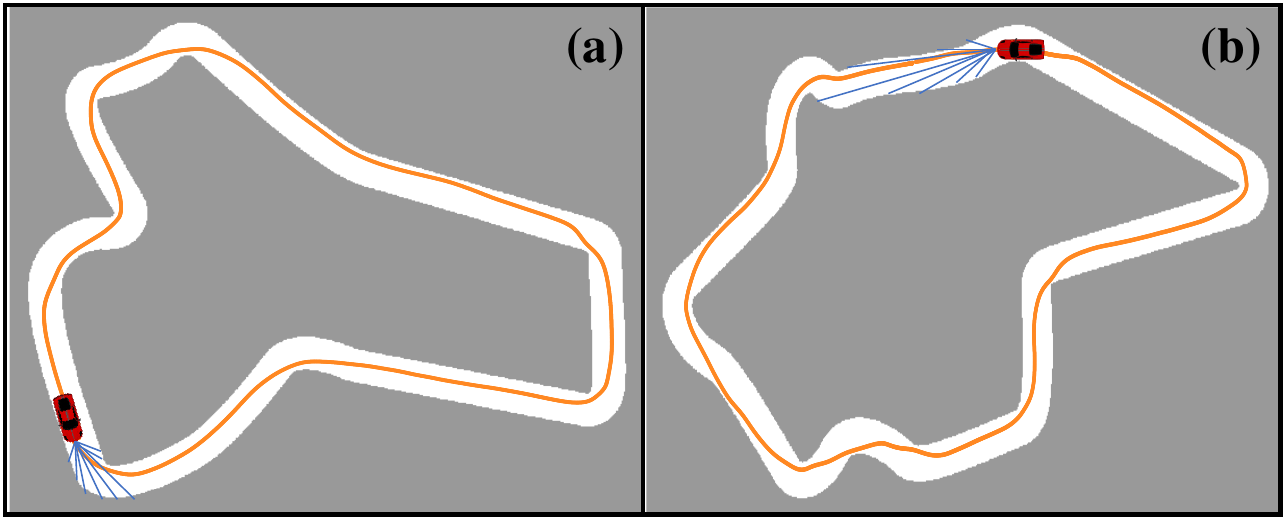}
    \caption{MPC-SafeGIL successfully completes laps on two maps: (a) the training map used for evaluation and comparison in simulation; (b) the unseen map used for testing the generalization capability of learned policies in Sec.~\ref {Sec:Generalization}. The \textcolor{orange}{orange} trajectories illustrate MPC-SafeGIL's successful navigation on both tracks.}
    \vspace{-1.0em}
    \label{fig:f110_map}
\end{figure}

To demonstrate the robustness of MPC-SafeGIL on visuomotor policy learning and generalization to out-of-distribution, we evaluate it on the F1Tenth racing car simulator. The policy takes high-dimensional LiDAR input to predict steering angle $s \in [-0.4189 \,, 0.4189] \mathrm{rad}$ and velocity $v \in [-5.0, 20.0] \mathrm{m/s}$. The task is to complete two laps on the track while avoiding collisions with curbs (Fig.~\ref{fig:f110_map}(a)). The failure set $\mathcal{L}$ includes all off-track states, and $l(x)$ is a signed distance function to the nearest track edge. The expert is a pure pursuit controller that determines steering angle and velocity using known map waypoints and has access to full state information during data collection. In contrast, the imitator operates using only LiDAR observations. The imitator is modeled as an MLP and trained for 10 different seeds to capture variance. Each policy is evaluated over 100 trajectories, with the car starting from a random position on the track. Evaluation metrics are distance traveled and collision rate.

To compute the optimal adversarial disturbance in MPC-SafeGIL, we use a disturbance bound of $\bar{d}_{max}=0.2\, \bar{u}$, with $N=500$ samples, a planning horizon of $H=100 (1s)$, and a smooth parameter $K=10$. These adversarial disturbances purposely steer the expert toward obstacle regions, creating safety-critical scenarios for demonstrations.

\begin{figure}[h]
    \centering
    \includegraphics[width=\linewidth]{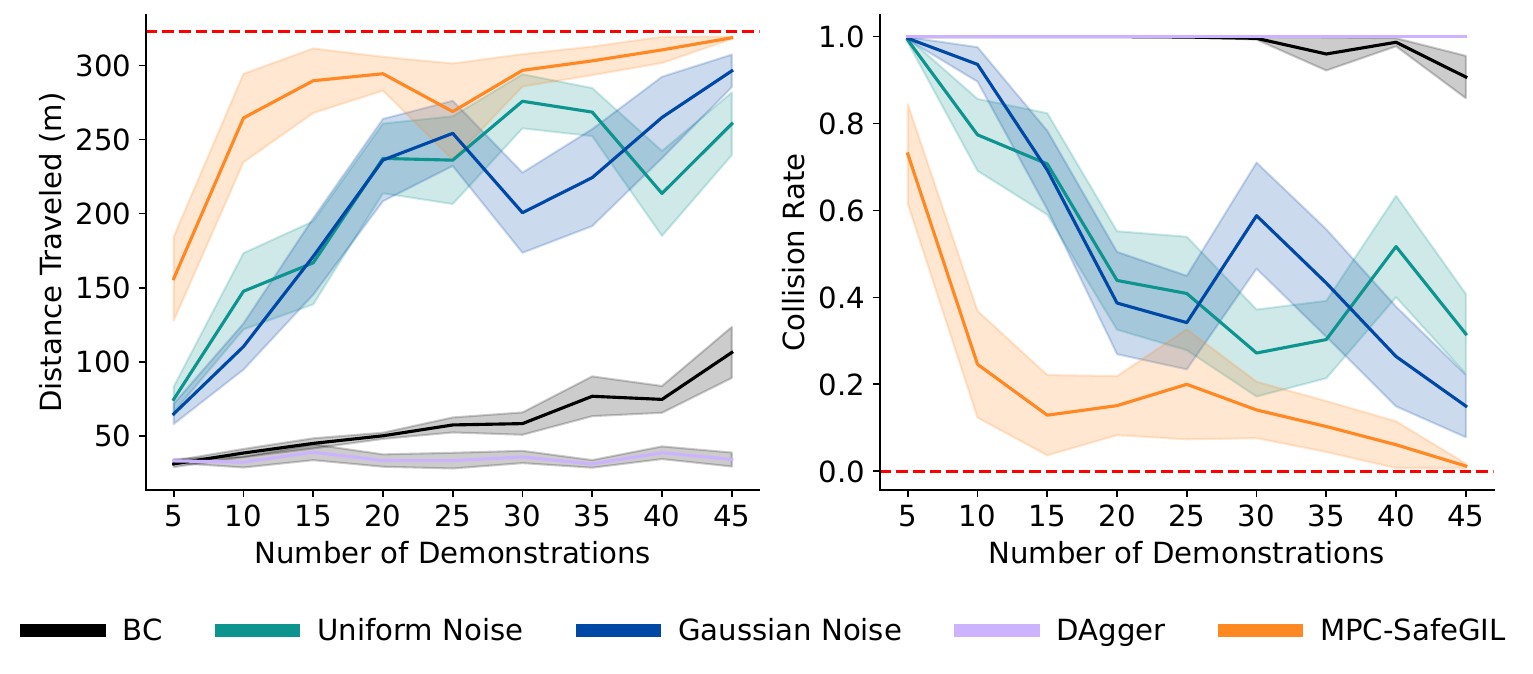}
    \caption{Comparison between MPC-SafeGIL and various baselines. The red dashed lines are expert performances. MPC-SafeGIL achieves the longest distance traveled and the lowest collision rate.}
    \vspace{-1.0em}
    \label{fig:f110_noise}
\end{figure}

\subsubsection*{\textbf{Comparison}} To assess the impact of adversarial noise, we compare MPC-SafeGIL against four baselines: vanilla BC, Gaussian noise, uniform noise and DAgger \cite{dagger}, an on-policy method that augments the training dataset with states visited by the learner. Fig.~\ref{fig:f110_noise} shows the results. DAgger suffers in this task due to the limitations of the scripted expert, which pursues the nearest waypoint. While the expert provides reasonable actions when the car closely follows its intended trajectory, it becomes unreliable and overly aggressive when queried in arbitrary individual states. This leads to a noisy and inconsistent training dataset, resulting in worse performance than BC. Injecting different types of noise improves both performance and safety compared to BC. In high-speed racing environments, even slight perturbations can help the car explore a wider range of states. However, MPC-SafeGIL consistently outperforms all baselines, achieving the longest distance traveled and the lowest collision rate. Notably, these gains are pronounced in low-data regimes, where MPC-SafeGIL exhibits faster and more significant improvements. These results highlight the importance of injecting structured, adversarial disturbances that guide the system into safety-critical states, enabling the imitation policy to learn more meaningful and robust recovery behaviors.

\begin{figure}[h]
    \centering
    \includegraphics[width=\linewidth]{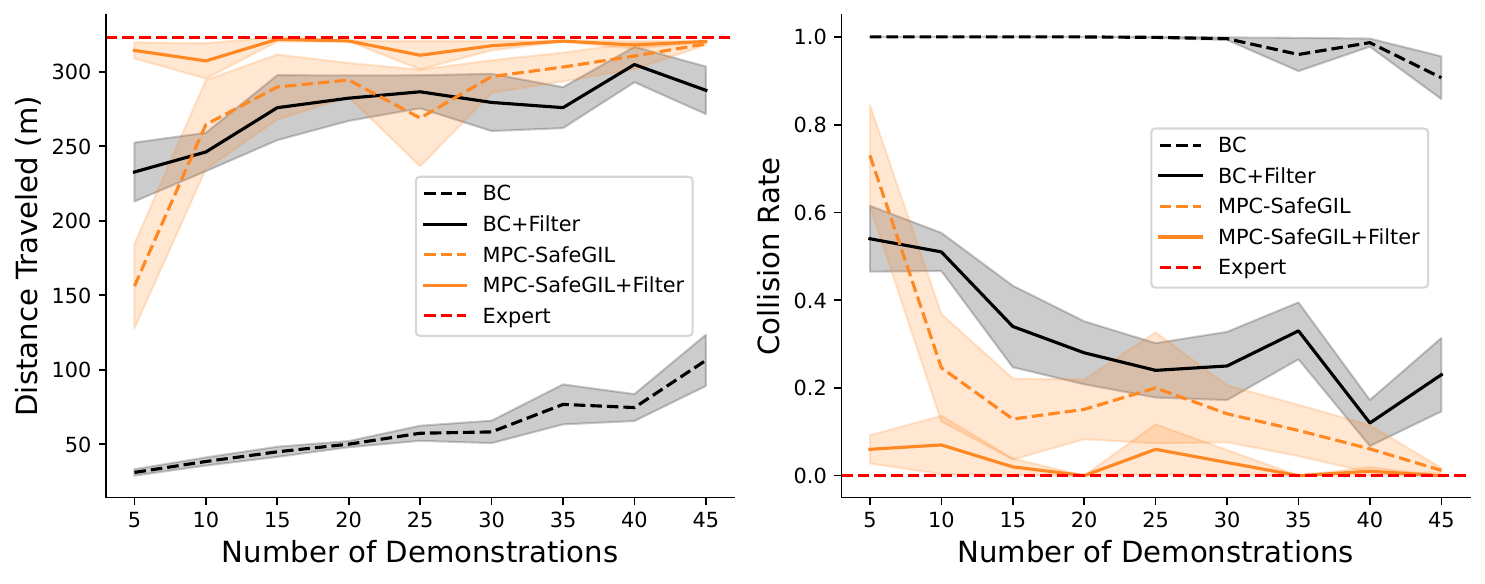}
    \caption{Evaluation of imitation policies with and without predictive safety filtering. The dashed lines represent evaluation for the imitation policy itself. The solid lines show the performance with predictive safety filtering.}
    \vspace{-1.0em}
    \label{fig:f110_filter}
\end{figure}

\subsubsection*{\textbf{Safety Filter}} To further demonstrate the seamless integration of the proposed approach with online safety mechanisms, we incorporate the imitation policies with a predictive safety filter at test time. The predictive safety filter overrides the nominal control with an MPC controller whenever a future $H=50$-step trajectory predicts a collision. As shown in Fig.~\ref{fig:f110_filter}, both BC and MPC-SafeGIL benefit from the safety filter, leading to a lower collision rate. However, MPC-SafeGIL consistently outperforms BC. Moreover, even without safety filter, MPC-SafeGIL achieves lower collision rates than BC with filter, showing its inherent safety-awareness. While the safety filter tries to select the safest control, it struggles to recover if the nominal policy places the car in a very dangerous position, such as heading directly toward curbs at high speed. These findings highlight the value of incorporating safety considerations during training, rather than relying solely on reactive corrections during testing.

\begin{table}[h]
    \centering
    \begin{tabular}{@{}c@{}c@{}c@{}c@{}c@{}}
\toprule
Method                  & BC            & MPC-SafeGIL        & Gaussian   & Uniform\\ \midrule
Distance(m)   & $25.55_{(1.13)}$   & $\mathbf{68.15}_{(6.54)}$   & $48.96_{(5.48)}$    & $42.88_{(9.50)}$ \\
\bottomrule
\end{tabular}

    \caption{Evaluation in an unseen environment.}
    \vspace{-1.0em}
    \label{tab:generalization}
\end{table}

\subsubsection*{\textbf{Generalization}} 
\label{Sec:Generalization}
To evaluate generalization, we test the best-performing policy from each method on a novel, unseen track (Fig.~\ref{fig:f110_map}(b)). The episode completes either upon completing one lap or colliding with curbs. The results in Table.~\ref{tab:generalization} show that MPC-SafeGIL achieves the highest average distance traveled of $68.15 \mathrm{m}$, outperforming BC and other noise-injected methods. By intentionally exposing the expert to diverse and safety-critical states during training, MPC-SafeGIL enables the learned policy to generalize more effectively to new environments. 
\vspace{-0.5em}

\subsection{Hardware Experiments}
We demonstrate our method on a resource-constrained quadrotor platform, Crazyflie 2.0, equipped with 8 time-of-flight (ToF) distance sensors and an onboard optical flow-based velocity estimator. The task is to navigate cluttered environments while avoiding collisions. Expert demonstrations are first collected in simulation, and the learned imitation policy is then deployed on the real-world platform. In simulation, the expert controller has access to privileged state and environment information. In contrast, the learned policy takes only the 8 ToF distance readings and planar velocity observations $(v_x, v_y)$ as input, and outputs desired roll and pitch commands $(\phi,\theta) \in [-0.1745, 0.1745]\mathrm{rad}$. The safety function $l(x)$ is defined as the signed distance from the quadrotor to the nearest obstacle. The failure set $\mathcal{L}$ thus consists of all states where the quadrotor collides with obstacles. The expert is an MPC-based controller using a 4D quadrotor dynamics model where $(p_x, p_y)$ denote position and $A_G$ denotes gravitational acceleration constant.
\begin{equation}
\begin{aligned}
    &\dot{p}_x = v_x,\quad &&\dot{p}_y = v_y,\\
    &\dot{v}_x = A_G \tan\theta,\quad &&\dot{v}_y = -A_G \tan\phi
\end{aligned}
\end{equation}
For each demonstration, the number of cylinder obstacles is uniformly sampled between $[1, 10]$; obstacle positions are sampled from $[0.5, 4.5] \times [0.5, 4.5] \mathrm{m}$ and their radius from $[0.1, 0.8]\mathrm{m}$. The quadrotor is initialized with a forward velocity of $v_x=1.0 \mathrm{m/s}$, starting from varying lateral positions along the $y$-axis. Each rollout terminates either upon collision or upon successfully reaching a $4 \mathrm{m}$ mark at the far end of the environment. To compute the optimal adversarial disturbance in MPC-SafeGIL, we use a disturbance bound of $\bar{d}_{max}=0.5\, \bar{u}$, with $N=1000$ samples, a planning horizon of $H=30(2s)$, and a smooth parameter $K=100$.

\subsubsection*{\textbf{Simulation Results}}
To evaluate policy performance in simulation, we conduct rollouts from 100 equally spaced initial positions along the $y$ axis in a controlled obstacle setting seen in Fig.~\ref{fig:quadrotor_data}. Each rollout ends when the quadrotor reaches the $4 \mathrm{m}$ goal mark without collision (success), when it crashes into an obstacle (fail), or when it fails to reach the mark within an episode limit (timeout). MPC-SafeGIL achieves a significantly higher safety rate of $73\%$ than BC, which achieves $41\%$.

\subsubsection*{\textbf{Sim2Real Gap}}
We further evaluate the real-world performance of the learned policies by deploying them on the Crazyflie platform in a physical environment replicating the simulated test obstacle configuration. The $y$-axis is partitioned into 12 equally spaced initial positions for testing. Once the quadrotor hovers from the initial position, the learned policy is activated. MPC-SafeGIL completed 9 out of 12 runs successfully without collisions, whereas the BC policy succeeded in only 3 runs. Notably, while the safety rate of BC drops from $41\%$ in simulation to $25\%$ in the real world, MPC-SafeGIL maintains consistent safety performance across both domains.

\begin{figure}[h]
    \centering
    \includegraphics[width=\linewidth]{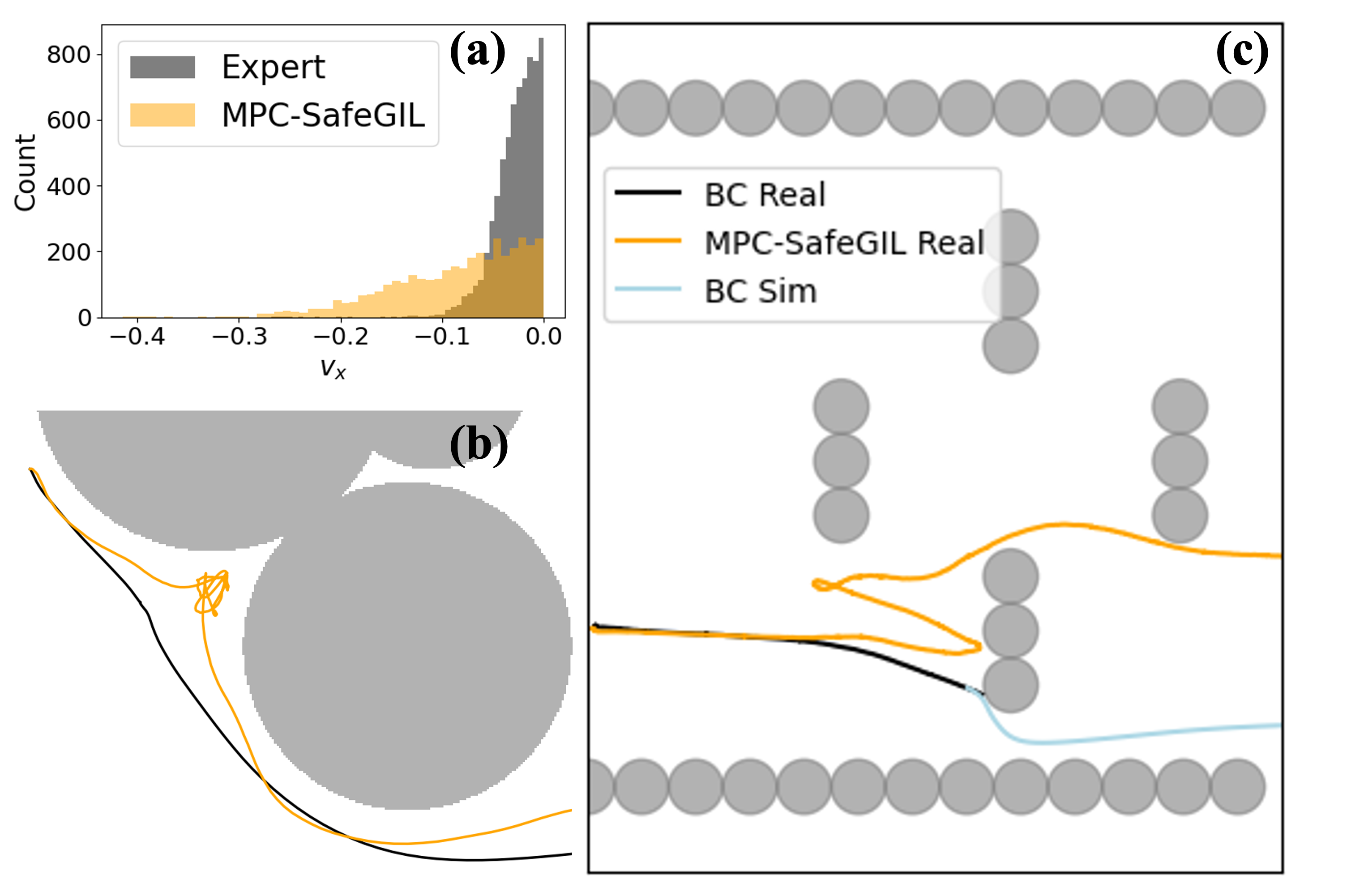}
    \caption{(a) The distribution of states with negative velocity in the training dataset. (b) While the normal expert demonstrates a direct path, safety guidance pushes the expert to demonstrate recovery maneuver. (c) Overlay of real-world recordings on simulation environment. Starting from a pre-crash state BC policy is able to succeed in simulation.}
    \vspace{-1.0em}
    \label{fig:quadrotor_data}
\end{figure}

The sim2real performance gap is primarily due to discrepancies in quadrotor dynamics, sensor noise, imperfect actuation and the difference of obstacle geometries. We hypothesize that these mismatches introduce additional uncertainty during real-world deployment, thereby increasing the likelihood of policy errors and potentially safety-critical situations. Fig.~\ref{fig:quadrotor_data} illustrates this qualitative difference in trajectories: while MPC-SafeGIL avoids obstacles in both domains, the BC policy succeeds in simulation but collides in the real world. To validate our hypothesis, we replay the BC policy from just before the real-world collision state in simulation and observe successful obstacle avoidance, suggesting that the failure arises from real-world discrepancies.

During real-world deployment, we observed a recurring failure mode in BC: when faced with an obstacle, the policy correctly retreats but fails to reattempt forward progress even when a clear opening appears, resulting in indefinite backward motion. Conversely, MPC-SafeGIL executes a robust recovery strategy--retreating, reorienting, and safely navigating through the opening, as shown in Fig.~\ref{fig:front figure}(a). This behavioral gap arises from differences in the state distributions encountered in the training dataset. Fig.~\ref{fig:quadrotor_data} shows that significant negative velocity states are underrepresented in the BC training dataset. Consequently, the learned BC policy generalizes poorly in such scenarios, continuing to move backward despite sensor readings indicating a clear path forward. In contrast, MPC-SafeGIL learns to recover from risky situations and complete the task effectively due to its exposure to a wider distribution of safety-critical states during training.

\subsubsection*{\textbf{Natural and Dynamic Environments}}
To further evaluate generalization capabilities, we deploy the policies in two challenging real-world scenarios: a densely cluttered environment and a dynamic scene with a walking pedestrian crossing the flight path (Fig.~\ref{fig:front figure}(b)(c)). In both cases, MPC-SafeGIL successfully avoids collisions and completes the navigation task, while the BC policy consistently fails. 
\vspace{-0.5em}

\section{Conclusion} \label{sec:conclusion}
We propose MPC-SafeGIL, a new algorithm to improve the safety of imitation learning in high-dimensional and black-box dynamical systems. We use MPC to compute optimal adversarial disturbances and inject them into expert demonstrations. This intentional exposure to safety-critical states allows the learned policy to capture more effective recovery behaviors. Experiments across various simulation studies and real-world hardware show that MPC-SafeGIL significantly enhances the safety of the learned policy without compromising task performance. These results highlight the potential of leveraging MPC to incorporate safety-awareness during the training phase of imitation learning.

\appendix[Reduction of Maxmin Problem]
\noindent We consider two optimal control problems with control-affine dynamics:

\noindent \textbf{Problem 1 (Maxmin Formulation):}
\begin{equation*}
    \begin{gathered}
    V_{1}(x, t) = \max_{u(\cdot)} \min_{d(\cdot)} \min_{k \in \{t,\dots, T\}} l(x_k) \\
    x_{k+1} =f_{1}(x_k)+f_{2}(u_k+d_k) \quad \forall k \in \{t,\dots, T\} \\
    |u_k| \leq \bar{u}, |d_k| \leq \bar{d}, \bar{u} > \bar{d}.
    \end{gathered}
\end{equation*}

\noindent \textbf{Problem 2 (Reduced Max Formulation):}
\[
\begin{aligned}
V_{2}(x, t) &= \max_{w(\cdot)} \min_{k \in \{t, \dots, T\} }l(x_k) \\
\text{s.t. } x_{k+1} = f_{1}(x_k) &+ f_{2}(x_k) w, \quad |w| \leq \bar{u} - \bar{d}.
\end{aligned}
\]

\noindent We now prove that $V_{1}(x, t) = V_{2}(x, t),\forall x, t.$ via induction over time. The dynamic programming principle for Problem~1 gives:
\begin{equation*}
    V_{1}(x_t, t) \approx \min \bigg\{l(x_t), \max_{u(\cdot)} \min_{d(\cdot)}V_{1}(x_{t+1}, t+1) \bigg\},
\end{equation*}
Applying first-order Taylor expansion:
\begin{equation*}
    V_{1}(x_{t+1}, t+1)  \approx V_{1}(x_t,t+1)
    + D_xV(x_t,t+1)(x_{t+1}-x_t),
\end{equation*}

\noindent Consider the control-affine dynamics and inner max-min:
\begin{align*}
    V_{1}&(x, T - 1) = \min \bigg \{l(x), \max_{u(\cdot)} \min_{d(\cdot)}V_{1}(x_T, T)  
    \bigg \} \\
    =&\min \bigg \{ l(x), \max_{u(\cdot)} \min_{d(\cdot)}V_{1}(x, T) + \\
    &D_xV_{1}(x, T) \cdot (f_{1}(x) + f_{2}(x)u + f_{2}(x)d -x) \bigg \} \\
    =&\min \bigg \{l(x),V_{1}(x,T)+ D_xV_{1}(x, T) \cdot (f_{1}(x) - x) \\&+ \left|D_xV_{1}(x,T) \cdot f_{2}(x) \right|\bar{u} - \left|D_xV_{1}(x,T) \cdot f_{2}(x) \right|\bar{d}\bigg \},
\end{align*}

\noindent Similarly, the dynamic programming update for $V_2$ gives:
\[
    \begin{aligned}
    V_{2}(x, T - 1) = & \min \bigg\{l(x), V_{2}(x, T) + D_xV_{2}(x, T) \cdot (f_{1}(x)-x) \\
    & + \left| D_xV_{2}(x,T) \cdot f_{2}(x)\right| \bar{w}\bigg\},
    \end{aligned}
    \label{eq:V2_Delta}
\]

\noindent Since $V_{1}(x, T) = V_{2}(x, T)=l(x)$ and $\bar{w}=\bar{u} - \bar{d}$, we have:
\[
V_{1}(x, T - 1) = V_{2}(x, T - 1), \quad \forall x
\]

\noindent Repeating this argument backward in time:
\[
V_{1}(x, t) = V_{2}(x, t), \quad \forall(x, t).
\]

\noindent Furthermore, if $w^{*}(x, t)$ is the optimal solution of Problem~2, then the optimal disturbance and control for Problem~1 are:
\[
d^{*}(x, t) =
\begin{cases}
-\bar{d}, & \text{if } w^{*}(x, t) > 0, \\
+\bar{d}, & \text{if } w^{*}(x, t) < 0.
\end{cases}
\]
\[
u^{*}(x, t) =
\begin{cases}
w^{*}(x, t) + \bar{d}, & \text{if } w^{*}(x, t) > 0, \\
w^{*}(x, t) - \bar{d}, & \text{if } w^{*}(x, t) < 0,
\end{cases}
\]

\vspace{-0.5em}
\bibliography{reference} 

\begin{thebibliography}{10}

\bibitem{finn2017one}
C.~Finn, T.~Yu, T.~Zhang, P.~Abbeel, and S.~Levine, ``One-shot visual imitation learning via meta-learning,'' in {\em CoRL}, PMLR, 2017.

\bibitem{zhao2023learning}
T.~Z. Zhao, V.~Kumar, S.~Levine, and C.~Finn, ``Learning fine-grained bimanual manipulation with low-cost hardware,'' {\em arXiv}, 2023.

\bibitem{bin2020learning}
X.~B. Peng, E.~Coumans, T.~Zhang, T.-W. Lee, J.~Tan, and S.~Levine, ``Learning agile robotic locomotion skills by imitating animals,'' {\em arXiv preprint arXiv:2004.00784}, 2020.

\bibitem{condinional_il}
F.~Codevilla, M.~M{\"u}ller, A.~Dosovitskiy, A.~M. L{\'o}pez, and V.~Koltun, ``End-to-end driving via conditional imitation learning,'' {\em IEEE ICRA}, 2018.

\bibitem{agile_auto_driving}
Y.~Pan, C.-A. Cheng, K.~Saigol, K.~Lee, X.~Yan, E.~A. Theodorou, and B.~Boots, ``Agile autonomous driving using end-to-end deep imitation learning,'' {\em Robotics: Science and Systems XIV}, 2017.

\bibitem{compounding_error_covariate_shift}
S.~Ross and D.~Bagnell, ``Efficient reductions for imitation learning,'' in {\em AISTATS}, 2010.

\bibitem{dagger}
S.~Ross, G.~Gordon, and D.~Bagnell, ``A reduction of imitation learning and structured prediction to no-regret online learning,'' in {\em AISTATS}, 2011.

\bibitem{choi2021robust}
J.~J. Choi, D.~Lee, K.~Sreenath, C.~J. Tomlin, and S.~L. Herbert, ``Robust control barrier--value functions for safety-critical control,'' in {\em IEEE CDC}, 2021.

\bibitem{wabersich2023data}
K.~P. Wabersich, A.~J. Taylor, J.~J. Choi, K.~Sreenath, C.~J. Tomlin, A.~D. Ames, and M.~N. Zeilinger, ``Data-driven safety filters: {H}amilton-{J}acobi reachability, control barrier functions, and predictive methods for uncertain systems,'' {\em IEEE Control Systems Magazine}, vol.~43, no.~5, pp.~137--177, 2023.

\bibitem{hsu2023safety}
K.-C. Hsu, H.~Hu, and J.~F. Fisac, ``The safety filter: A unified view of safety-critical control in autonomous systems,'' {\em arXiv preprint arXiv:2309.05837}, 2023.

\bibitem{yang2024enhancing}
Y.~Yang, L.~Chen, Z.~Zaidi, S.~van Waveren, A.~Krishna, and M.~Gombolay, ``Enhancing safety in learning from demonstration algorithms via control barrier function shielding,'' in {\em ACM/IEEE HRI}, 2024.

\bibitem{ames2019control}
A.~D. Ames, S.~Coogan, M.~Egerstedt, G.~Notomista, K.~Sreenath, and P.~Tabuada, ``Control barrier functions: Theory and applications,'' in {\em IEEE ECC}, 2019.

\bibitem{prajna2004safety}
S.~Prajna and A.~Jadbabaie, ``Safety verification of hybrid systems using barrier certificates,'' in {\em International Workshop on Hybrid Systems: Computation and Control}, 2004.

\bibitem{bajcsy2019efficient}
A.~Bajcsy, S.~Bansal, E.~Bronstein, V.~Tolani, and C.~J. Tomlin, ``An efficient reachability-based framework for provably safe autonomous navigation in unknown environments,'' in {\em IEEE CDC}, 2019.

\bibitem{nguyen2024gameplay}
D.~P. Nguyen, K.-C. Hsu, W.~Yu, J.~Tan, and J.~F. Fisac, ``Gameplay filters: Robust zero-shot safety through adversarial imagination,'' {\em arXiv preprint arXiv:2405.00846}, 2024.

\bibitem{ke2023ccil}
L.~Ke, Y.~Zhang, A.~Deshpande, S.~Srinivasa, and A.~Gupta, ``{CCIL: C}ontinuity-based data augmentation for corrective imitation learning,'' in {\em International Conference on Learning Representations}, 2024.

\bibitem{zhou2023nerf}
A.~Zhou, M.~J. Kim, L.~Wang, P.~Florence, and C.~Finn, ``Nerf in the palm of your hand: {C}orrective augmentation for robotics via novel-view synthesis,'' in {\em IEEE/CVF CVPR}, 2023.

\bibitem{hoque2024intervengen}
R.~Hoque, A.~Mandlekar, C.~Garrett, K.~Goldberg, and D.~Fox, ``Intervengen: Interventional data generation for robust and data-efficient robot imitation learning,'' {\em arXiv preprint arXiv:2405.01472}, 2024.

\bibitem{laskey2017dart}
M.~Laskey, J.~Lee, R.~Fox, A.~Dragan, and K.~Goldberg, ``{DART}: {N}oise injection for robust imitation learning,'' in {\em CoRL}, 2017.

\bibitem{gail}
J.~Ho and S.~Ermon, ``Generative adversarial imitation learning,'' in {\em Neurips}, 2016.

\bibitem{airl}
J.~Fu, K.~Luo, and S.~Levine, ``Learning robust rewards with adversarial inverse reinforcement learning,'' {\em ArXiv}, vol.~abs/1710.11248, 2017.

\bibitem{sqil}
S.~Reddy, A.~D. Dragan, and S.~Levine, ``Sqil: Imitation learning via reinforcement learning with sparse rewards,'' {\em arXiv: Learning}, 2019.

\bibitem{tasil}
D.~Pfrommer, T.~Zhang, S.~Tu, and N.~Matni, ``Tasil: Taylor series imitation learning,'' {\em ArXiv}, vol.~abs/2205.14812, 2022.

\bibitem{stableBC}
S.~A. Mehta, Y.~U. Ciftci, B.~Ramachandran, S.~Bansal, and D.~P. Losey, ``Stable-bc: Controlling covariate shift with stable behavior cloning,'' {\em arXiv preprint arXiv:2408.06246}, 2024.

\bibitem{ciftci2024safe}
Y.~U. Ciftci, D.~Chiu, Z.~Feng, G.~S. Sukhatme, and S.~Bansal, ``Safe-gil: Safety guided imitation learning for robotic systems,'' {\em ICRA}, 2024.

\bibitem{bansal2017hamilton}
S.~Bansal, M.~Chen, S.~Herbert, and C.~J. Tomlin, ``{Hamilton-Jacobi Reachability}: A brief overview and recent advances,'' in {\em IEEE CDC}, 2017.

\bibitem{mitchell2005time}
I.~M. Mitchell, A.~M. Bayen, and C.~J. Tomlin, ``A time-dependent hamilton-jacobi formulation of reachable sets for continuous dynamic games,'' {\em IEEE Transactions on automatic control}, vol.~50, no.~7, pp.~947--957, 2005.

\bibitem{mitchell2004toolbox}
I.~Mitchell, ``A toolbox of level set methods,'' {\em http://www. cs. ubc. ca/mitchell/ToolboxLS/toolboxLS. pdf, Tech. Rep. TR-2004-09}, 2004.

\bibitem{aggressivemppi}
G.~Williams, P.~Drews, B.~Goldfain, J.~M. Rehg, and E.~A. Theodorou, ``Aggressive driving with model predictive path integral control,'' in {\em IEEE ICRA}, 2016.

\bibitem{Borquez2025DualGuard}
J.~Borquez, L.~Raus, Y.~U. Ciftci, and S.~Bansal, ``Dualguard mppi: Safe and performant optimal control by combining sampling-based mpc and hamilton-jacobi reachability,'' {\em IEEE RAL}, 2025.

\bibitem{multirotor}
J.~Pravitra, K.~A. Ackerman, C.~Cao, N.~Hovakimyan, and E.~A. Theodorou, ``$\mathcal{L}$ 1-adaptive mppi architecture for robust and agile control of multirotors,'' in {\em 2020 IEEE/RSJ IROS}, 2020.

\bibitem{lynch2017modern}
K.~M. Lynch and F.~C. Park, {\em Modern robotics}.
\newblock Cambridge University Press, 2017.

\bibitem{makoviychuk2021isaac}
V.~Makoviychuk, L.~Wawrzyniak, Y.~Guo, M.~Lu, K.~Storey, M.~Macklin, D.~Hoeller, N.~Rudin, A.~Allshire, A.~Handa, {\em et~al.}, ``Isaac gym: High performance gpu-based physics simulation for robot learning,'' {\em arXiv preprint arXiv:2108.10470}, 2021.

\bibitem{margolis2022walktheseways}
G.~B. Margolis and P.~Agrawal, ``Walk these ways: Tuning robot control for generalization with multiplicity of behavior,'' {\em CoRL}, 2022.

\end{thebibliography}
\bibliographystyle{ieeetr}
\end{document}